\documentclass[letterpaper, 10 pt, conference]{ieeeconf} 

\IEEEoverridecommandlockouts                          
\usepackage{graphics} 
\usepackage{epsfig} 
\usepackage{mathptmx} 
\usepackage{times} 
\usepackage{amsmath} 
\usepackage{amssymb} 
\usepackage{hyperref}
\usepackage{subcaption}
\usepackage{xcolor}
\usepackage{tabularx}
\usepackage{afterpage}
\usepackage{url}

\title{\LARGE \bf
	ExACT: An End-to-End Autonomous Excavator System \\ Using Action Chunking With Transformers
}

\author{Liangliang Chen, Shiyu Jin, Haoyu Wang, Liangjun Zhang\\
	\thanks{Robotics and Autonomous Driving Lab, Baidu Research, USA. liangjunzhang@baidu.com. This work was done when Liangliang Chen was an intern at Baidu Research.
}}

\begin{document}
	\maketitle
	\thispagestyle{empty}
	\pagestyle{empty}
	
	\begin{abstract}
		Excavators are crucial for diverse tasks such as construction and mining, while autonomous excavator systems enhance safety and efficiency, address labor shortages, and improve human working conditions. Different from the existing modularized approaches, this paper introduces ExACT, an end-to-end autonomous excavator system that processes raw LiDAR, camera data, and joint positions to control excavator valves directly. Utilizing the Action Chunking with Transformers (ACT) architecture, ExACT employs imitation learning to take observations from multi-modal sensors as inputs and generate actionable sequences. In our experiment, we build a simulator based on the captured real-world data to model the relations between excavator valve states and joint velocities. With a few human-operated demonstration data trajectories, ExACT demonstrates the capability of completing different excavation tasks, including reaching, digging and dumping through imitation learning in validations with the simulator. To the best of our knowledge, ExACT represents the first instance towards building an end-to-end autonomous excavator system via imitation learning methods with a minimal set of human demonstrations. The video about this work can be accessed at  \href{https://youtu.be/NmzR_Rf-aEk}{https://youtu.be/NmzR\_Rf-aEk}.
	\end{abstract}
	
	\section{Introduction}
	Excavators are highly versatile heavy equipment, crucial for applications ranging from mining and construction to environmental restoration and emergency rescue, capable of operating in diverse and challenging conditions \cite{johns2023framework}. Autonomous excavators offer a critical solution to the dangers and inefficiencies of traditional excavation \cite{eraliev2022sensing}. By operating independently, these machines significantly enhance safety, mitigating the high number of injuries and fatalities due to hazardous excavation environments. They are especially beneficial in remote and harsh locations, where extreme conditions and a scarcity of skilled operators challenge productivity. Autonomous systems also circumvent human limitations, such as fatigue, and address workforce issues like aging and labor shortages \cite{zhang2021autonomous}. To sum up, autonomous excavators promise to not only improve safety and operational throughput but also provide a sustainable response to labor challenges in the construction and mining industries. 
	
	The existing autonomous excavator systems execute modular tasks in sequential order, i.e., perception, prediction, planning, and control \cite{eraliev2022sensing, zhang2021autonomous}. Zhang \textit{et al.} \cite{zhang2021autonomous} designed an autonomous excavator system for material handling in challenging environments such as mining and construction, utilizing a robust architecture that integrates multimodal perception sensors like LiDAR and cameras with advanced image processing and task planning algorithms. However, the hydraulic excavator is a highly nonlinear system in which the effects of working load, control input delay, and dead zone are hard to capture by the physical models in modular methods. 
	\begin{figure}[!t]
		\includegraphics[width=0.49\textwidth]{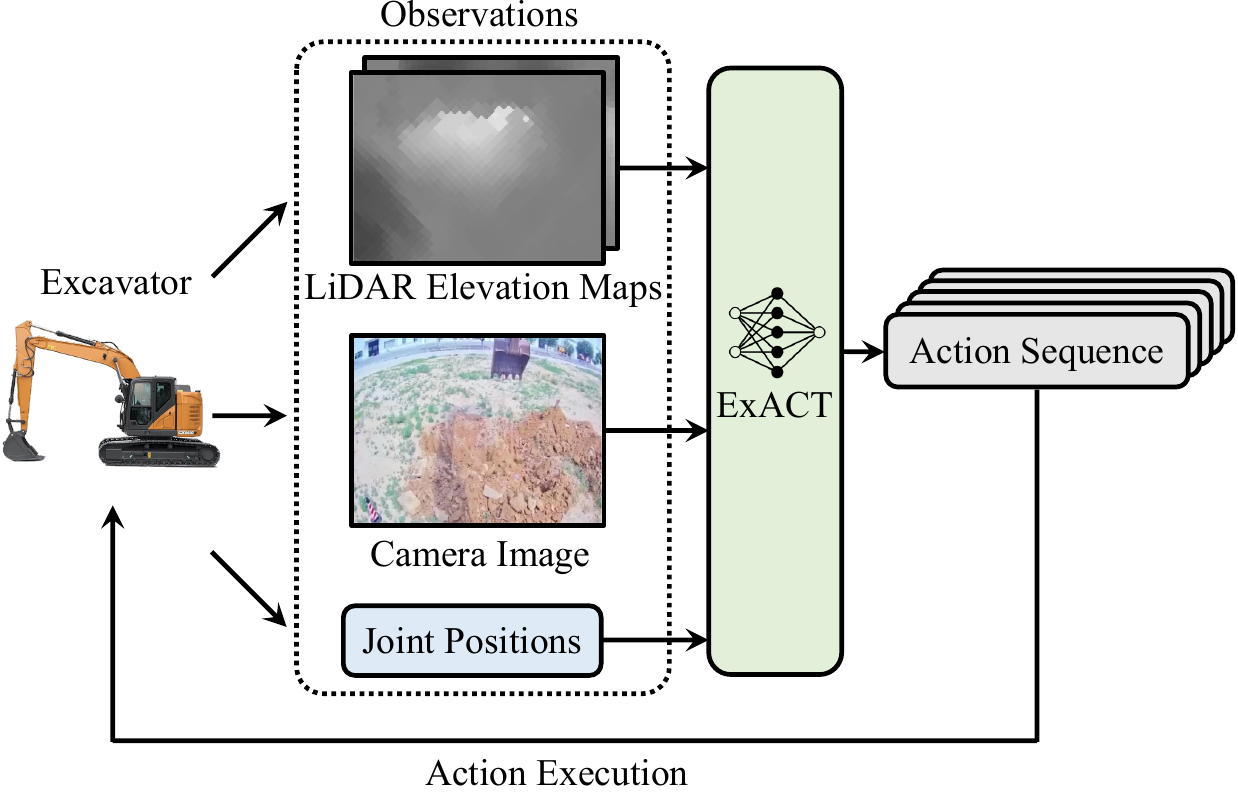}
		\caption{Framework of ExACT}
		\label{Fig7}
	\end{figure}
	Data-driven methods overcome this issue by directly deriving a model or controller from collected data without physical modeling \cite{hou2013model}. Many works have investigated the data-driven modeling or control strategies for excavator systems \cite{sandzimier2020data, lee2022precision}. The learning-based strategy is an important type of data-driven method that is being actively explored in the field of autonomous excavator systems. Ref. \cite{egli2022general} investigated using reinforcement learning to obtain end-effector trajectory tracking controllers for hydraulic excavators. Jin {\textit{et al.}} \cite{jin2023learning} developed an offline reinforcement learning excavator controller based on implicit Q-learning \cite{kostrikov2021offline}, which does not require online interactions between the controller and environment. 
	
	Recently, efficient imitation learning-based methods have been developed for robot manipulation \cite{chi2023diffusion, zhao2023learning}. Ref. \cite{chi2023diffusion} developed the diffusion policy, a robot imitation learning algorithm based on the diffusion model \cite{ho2020denoising}. However, the inference time of the diffusion policy may be too long to satisfy the requirement of high-frequency excavator control. Zhao {\textit{et al.}} \cite{zhao2023learning} proposed another imitation learning method ACT based on the conditional variational autoencoder \cite{kingma2013auto, sohn2015learning} and the transformer architecture \cite{vaswani2017attention}. 
	
	\begin{figure*}
		\centering
		\scalebox{0.95}{
			\begin{subfigure}[c]{0.347\textwidth}
				\centering
				\includegraphics[width=\textwidth]{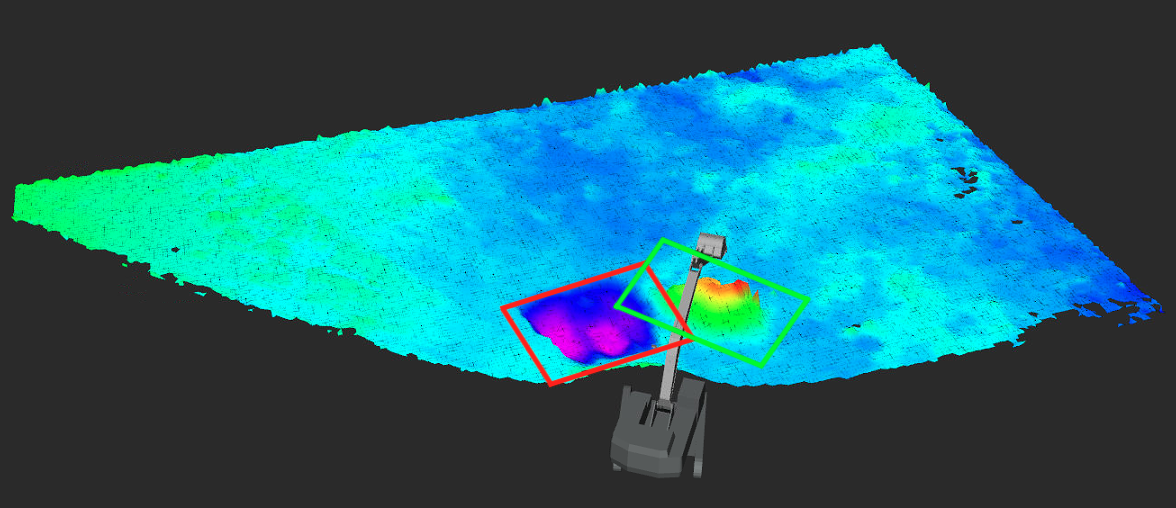}
				{\small(a)}
			\end{subfigure}
			\begin{subfigure}[c]{0.2\textwidth}
				\centering
				\includegraphics[width=\textwidth]{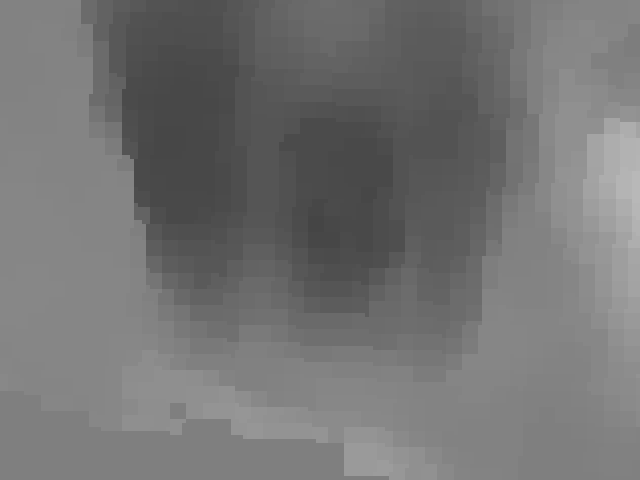}
				{\small(b)}
			\end{subfigure}
			\begin{subfigure}[c]{0.2\textwidth}
				\centering
				\includegraphics[width=\textwidth]{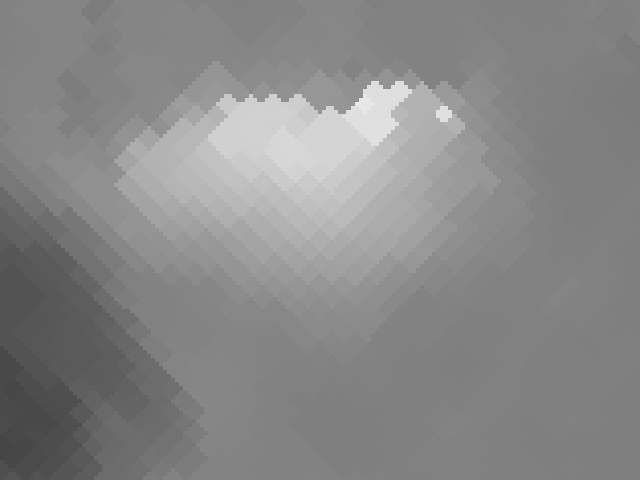}
				{\small(c)}
			\end{subfigure}
			\begin{subfigure}[c]{0.2\textwidth}
				\centering
				\includegraphics[width=\textwidth]{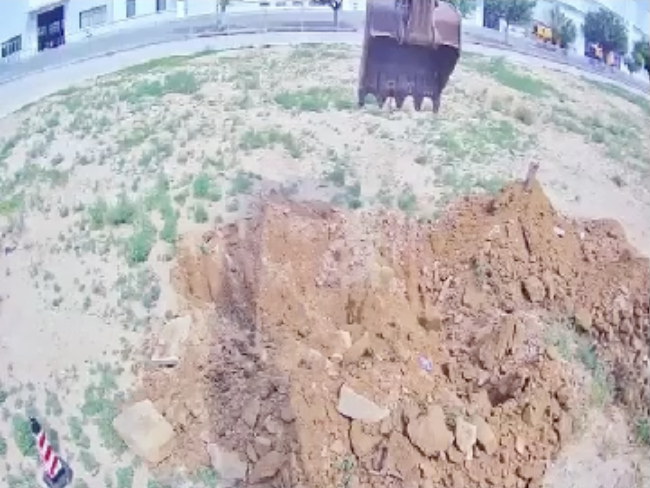}
				{\small(d)}
			\end{subfigure}
		}
		\caption{Examples of the front camera image and LiDAR elevation maps. (a) Raw LiDAR elevation map (visualized in RViz) from which the digging (red box) and dumping (green box) LiDAR elevation maps are cropped; (b) Preprocessed LiDAR elevation map of the digging zone; (c) Preprocessed LiDAR elevation map of the dumping zone; (d) front camera image.}
		\label{Fig0}
	\end{figure*}
	
	This paper proposes ExACT, an end-to-end autonomous excavator system, which takes the raw LiDAR and camera inputs and directly outputs valve commands to control the excavator. Our system leverages ACT architecture for imitation learning. Utilizing observations from multi-modal sensors such as LiDAR, cameras, and inclination sensors, ExACT generates a sequence of actions that can be executed sequentially. With only a limited number of human demonstration data collected from a real excavator, we show that the excavator can complete a set of tasks through imitation learning with only a few comprehensive demonstration trajectories. To the best of our knowledge, this is the first instance towards building an end-to-end autonomous excavator system via imitation learning methods with a minimal set of human demonstrations. 
	
	\section{ExACT: An End-to-End Autonomous Excavator System}
	The proposed autonomous excavator system, ExACT, leverages ACT to train an excavator controller by imitation learning. ACT is an imitation learning algorithm that has demonstrated excellent performance on end-to-end bimanual robotic manipulations \cite{zhao2023learning, fu2024mobile}. Different from the traditional imitation learning methods, such as behavioral cloning \cite{pomerleau1988alvinn}, ACT addresses the compounding errors by using action chunking, i.e., predicting a sequence of actions, and temporal ensembling to generate smooth actions. In addition, ACT leverages a conditional variational autoencoder \cite{sohn2015learning} at a high level to generate action sequences. Employing a generative model strengthens the controller's robustness against the noise in human demonstration data.
	
	The method framework is presented in Fig. \ref{Fig7}. The controller inputs consist of a camera image and the excavator joint positions. Unlike the original ACT algorithm \cite{zhao2023learning}, ExACT considers the LiDAR elevation maps of the digging and dumping areas as additional inputs if the excavator task involves digging and dumping. The controller outputs a sequence of actions that can be implemented on the excavator to fulfill the task. We use temporal ensembling to make the actions smooth. With this strategy, we make an action sequence prediction at each timestep and ensemble the relevant actions by the exponential weighting scheme. The obtained ensembled actions are implemented in the excavator. The experiment details can be found in Section \ref{S3}. 
	
	For the digging and dumping tasks, the 3D geometry information of the digging and dumping zones is important for the excavator controller to decide where to dig and dump. The LiDAR equipment detects the elevations of the workspace, from which we can extract the elevation maps of the digging and dumping zones. We cropped two local elevation maps from the raw LiDAR elevation map that provides ExACT with accurate 3D geometry information of the digging and dumping zones. 
	
    \section{Experiments}
    \label{S3} 
	In the experiments, the robotic platform employs a 21.5-ton hydraulic excavator that has been upgraded with a drive-by-wire system. This allows the excavator to be operated either through a handheld remote control (RC) or controlled via computer using a CAN bus interface. The excavator is further equipped with a variety of sensors, including inclinometers for measuring the joint pose of the excavator's boom, stick, and bucket, an encoder for measuring the swing angle of the cabin, a 3D LiDAR sensor, multiple RGB cameras, an IMU, and a real-time kinematic (RTK) GPS. The excavator is installed with an industry-standard computer with a GPU.
	
	\subsection{Excavator Task Descriptions}
	\label{S31}
	The following excavator tasks are considered in our experiments. 
	\begin{itemize}
		\item [1)] \texttt{reach}: Use the excavator bucket to reach a fixed target position, indicated by a traffic cone, from different locations. The controlled variables of this task are the valve states, which means that this is an end-to-end control task. 
		\item [2)] \texttt{dig\_dump}: Dig soil in a certain area and dump it in another area. We consider the valve states as the controlled variables for this task.
		\item [3)] \texttt{dig\_dump\_return}: Dig soil in a certain area, dump it to another area, swing the cabin, and return the bucket to a position above the digging area for the next circle of digging. The controlled variables for this task are the joint positions. 
	\end{itemize}
	
	The input variables of all the tasks above include: i) the 4-dimensional joint positions (i.e., swing, boom, stick, and bucket); ii) a front camera image with the size $480\times640\times3$; iii) the LiDAR elevation maps of the digging and dumping zones with the sizes $480\times640\times3$, respectively, if the task involves digging and dumping. The examples of the LiDAR elevation maps and the front camera image are shown in Fig. \ref{Fig0}. 
	
	\begin{figure*}[!t]
		\begin{minipage}[b]{0.7\textwidth}
            \centering
            \hspace{-2.0cm}
            \vspace{0.0cm}
		\scalebox{1.0}{
                    \includegraphics[width=0.2\textwidth]
                {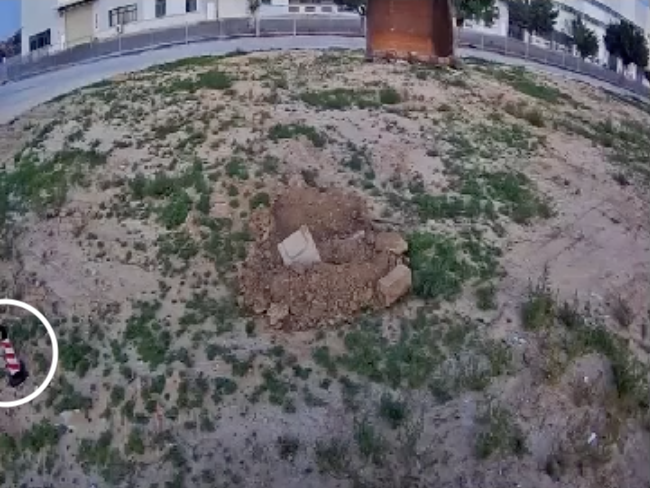}
				\includegraphics[width=0.2\textwidth]{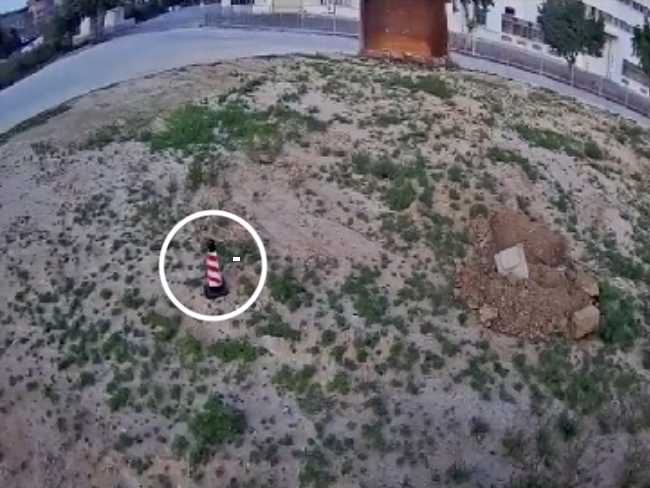}
				\includegraphics[width=0.2\textwidth]{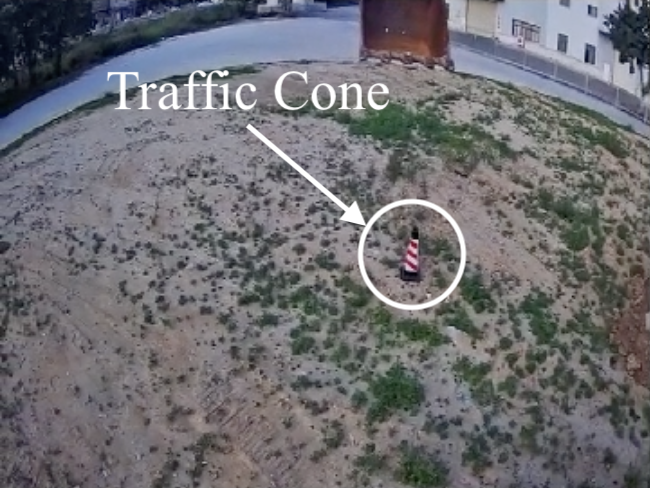}
				\includegraphics[width=0.2\textwidth]{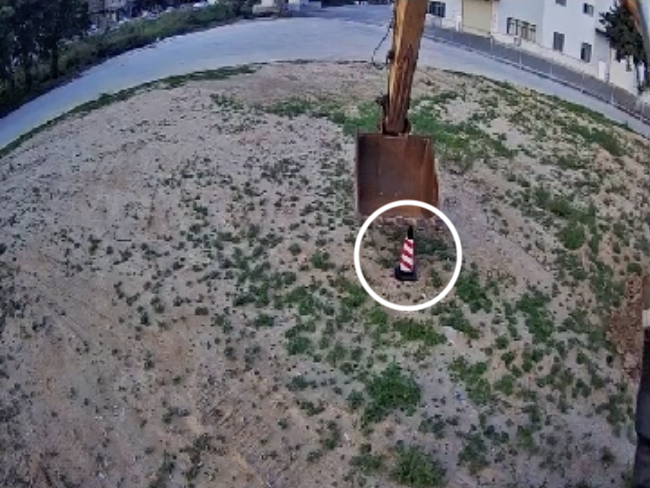}
		}
  
		{\hspace{-2.0cm}\small(a) Front camera images\vspace{0.1cm}}
  
		\hspace{-2.0cm}
		\scalebox{1.0}{
				\includegraphics[width=0.2\textwidth]{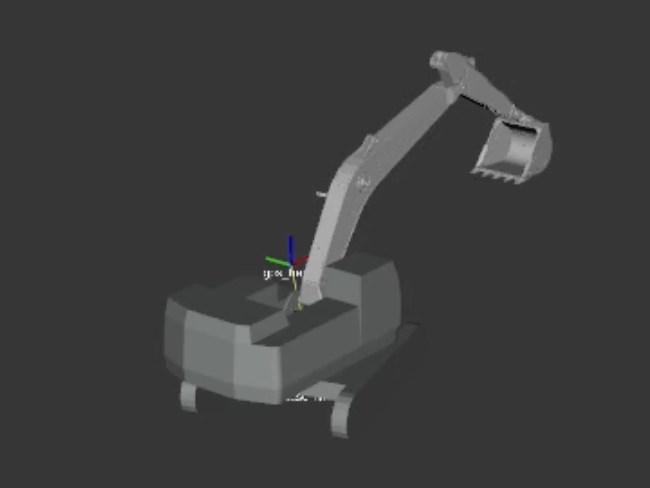}
				\includegraphics[width=0.2\textwidth]{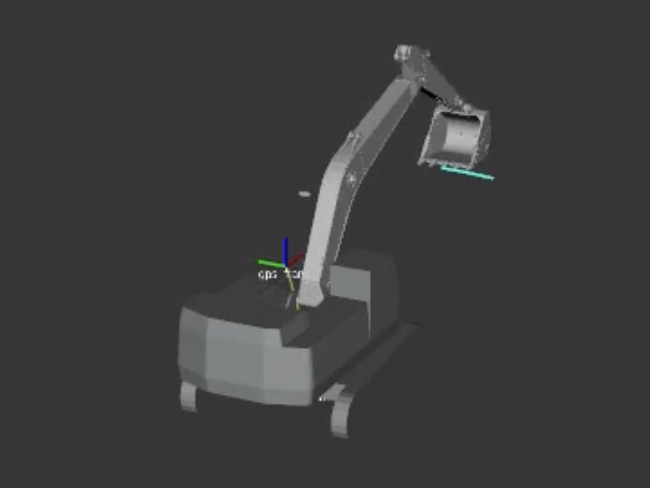}
				\includegraphics[width=0.2\textwidth]{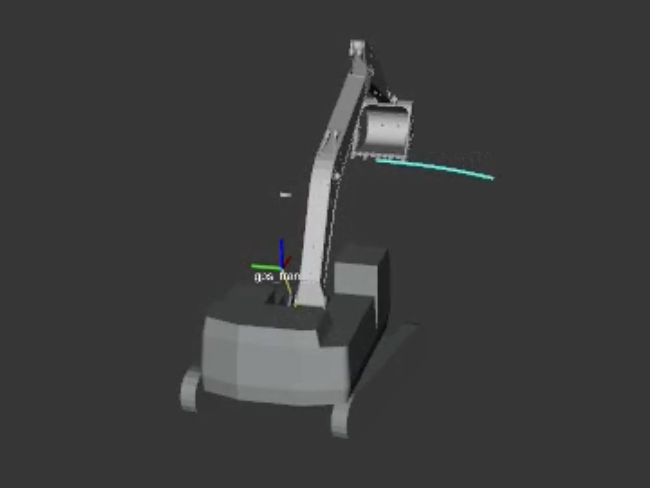}
				\includegraphics[width=0.2\textwidth]{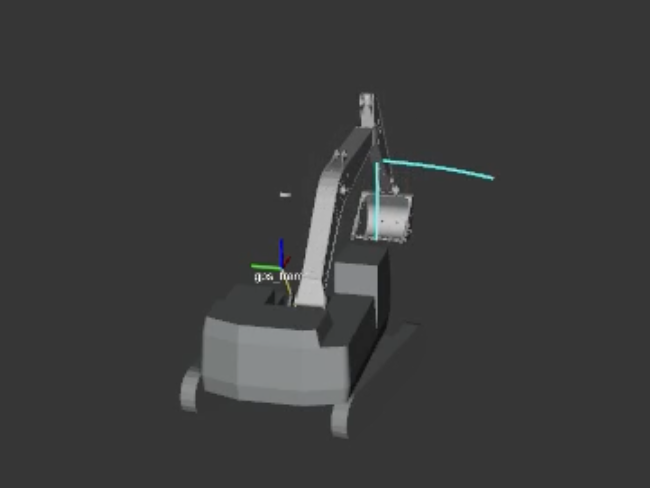}
		}
		
		{\hspace{-2.0cm}\vspace{0.cm}\small(b) Bucket trajectory visualizations in RViz\vspace{0.5cm}}
        \end{minipage}
        \hfill\hspace{-1.5cm}
        \begin{minipage}[b]{0.38\textwidth}
        \centering
        \includegraphics[width=\textwidth]{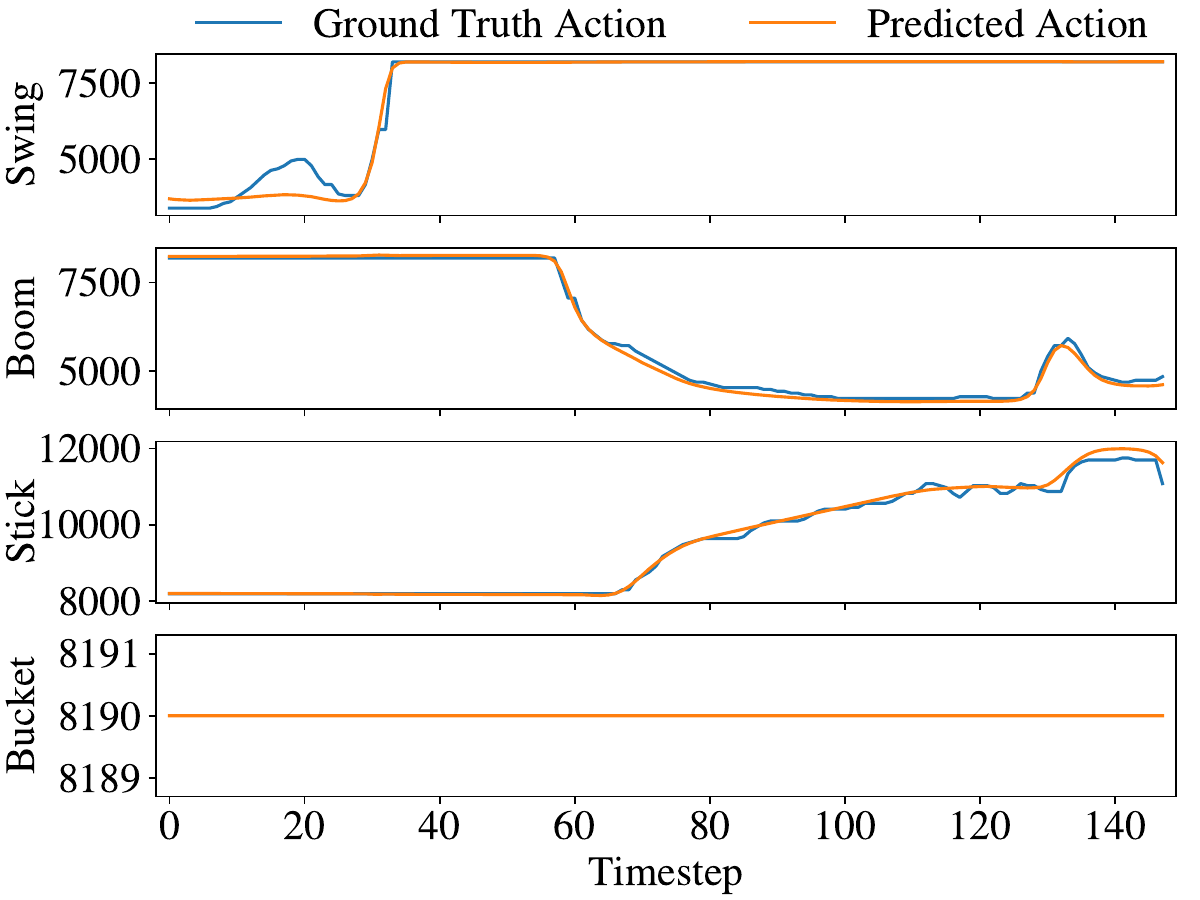}
        {\small (c) Ground truth and predicted actions}
        \end{minipage}
		\caption{Test performance of the task \texttt{reach} (valve state control)}
		\label{Fig1}
	\end{figure*}

        \begin{figure}[!t]
		\includegraphics[width=0.49\textwidth]{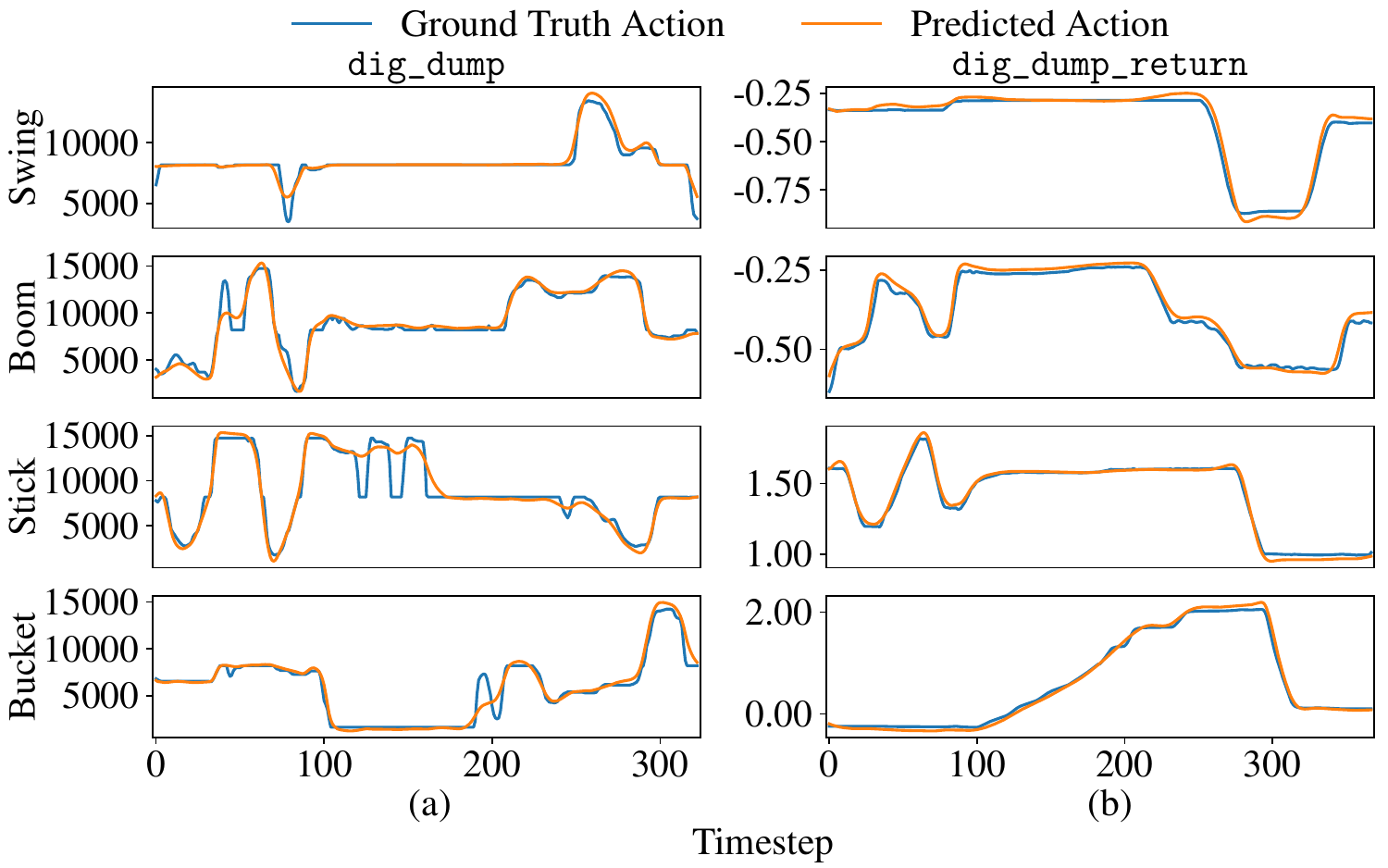}
		\caption{Ground truth and predicted actions of the tasks (a) \texttt{dig\_dump} and (b) \texttt{dig\_dump\_return}}
		\label{Fig6}
	\end{figure}
	
	\begin{figure*}[htbp]
		\centering
		\scalebox{0.95}{
			\begin{minipage}{0.1\textwidth}
				\includegraphics[width=\textwidth]{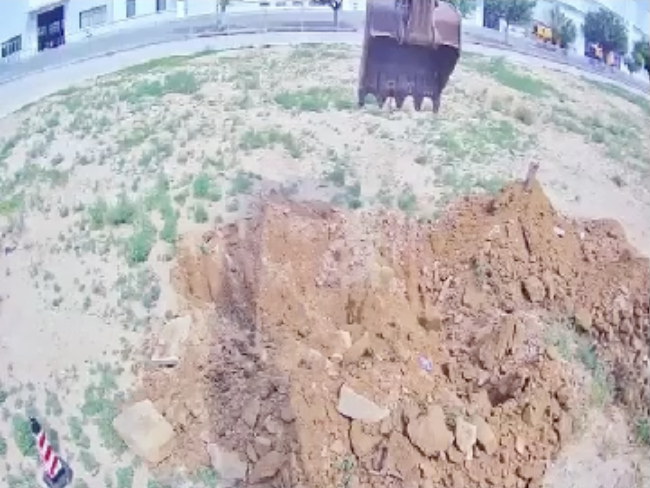}
			\end{minipage}
                \hspace{-0.16cm}
			\begin{minipage}{0.1\textwidth}
				\includegraphics[width=\textwidth]{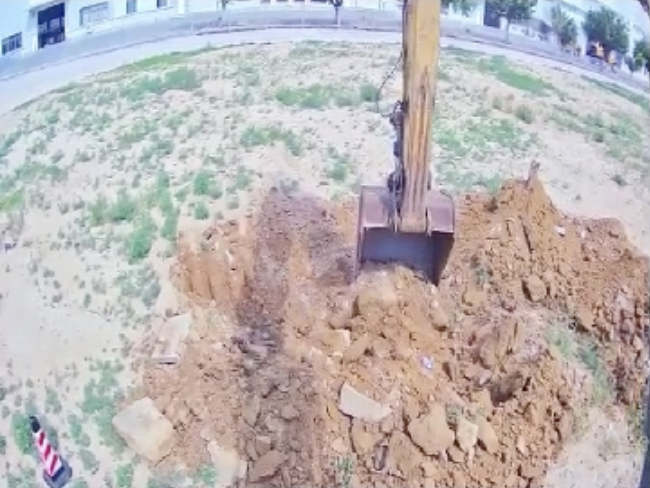}
			\end{minipage}
                \hspace{-0.16cm}
			\begin{minipage}{0.1\textwidth}
				\includegraphics[width=\textwidth]{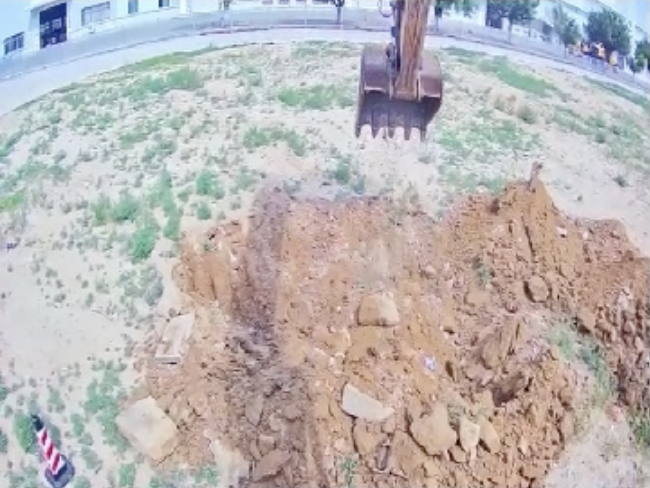}
			\end{minipage}
                \hspace{-0.16cm}
			\begin{minipage}{0.1\textwidth}
				\includegraphics[width=\textwidth]{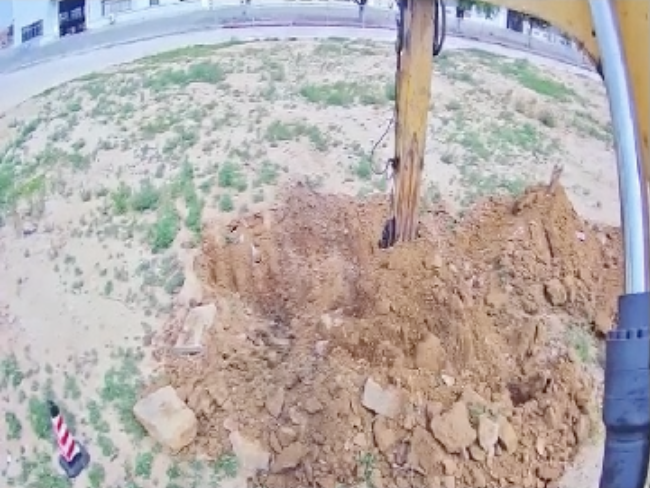}
			\end{minipage}
                \hspace{-0.16cm}
			\begin{minipage}{0.1\textwidth}
				\includegraphics[width=\textwidth]{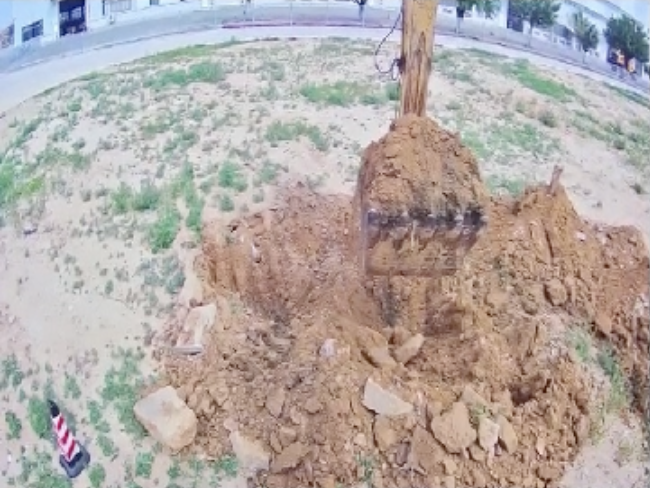}
			\end{minipage}
                \hspace{-0.16cm}
                \begin{minipage}{0.1\textwidth}
				\includegraphics[width=\textwidth]{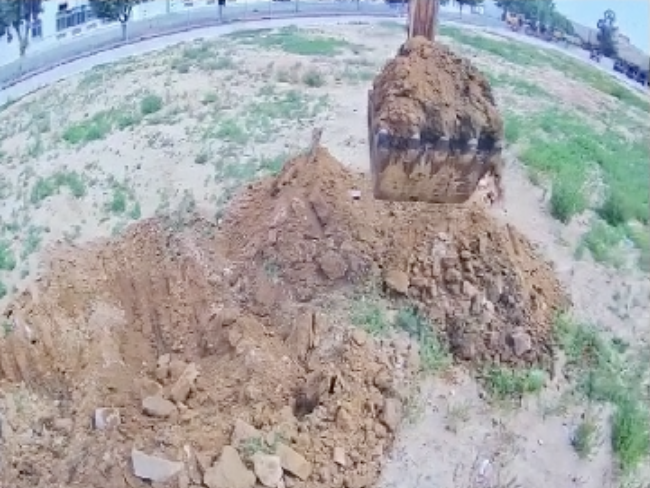}
			\end{minipage}
                \hspace{-0.16cm}
			\begin{minipage}{0.1\textwidth}
				\includegraphics[width=\textwidth]{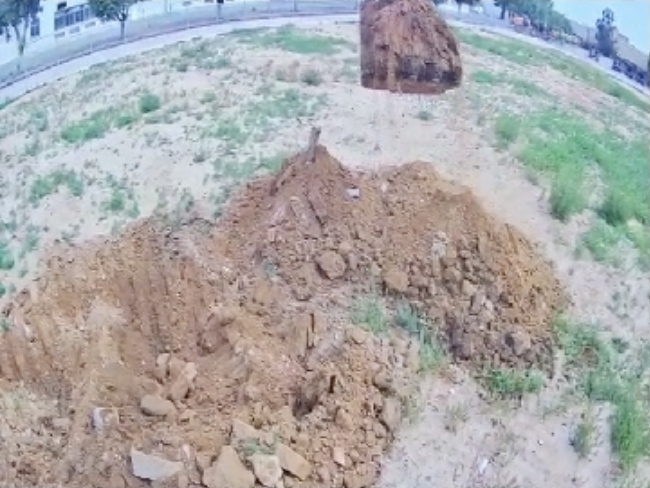}
			\end{minipage}
                \hspace{-0.16cm}
			\begin{minipage}{0.1\textwidth}
				\includegraphics[width=\textwidth]{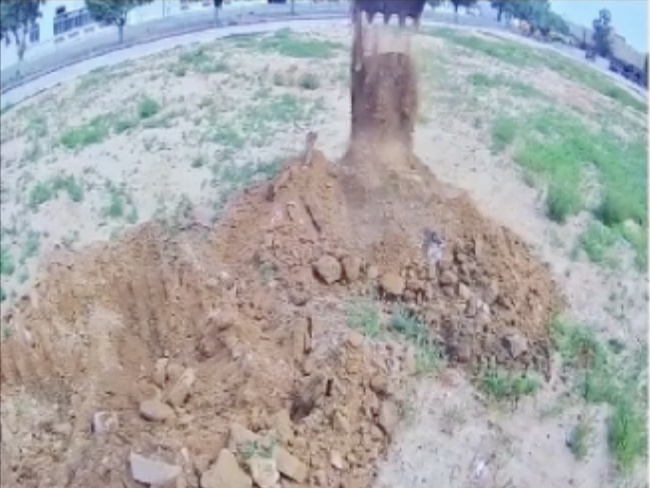}
			\end{minipage}
                \hspace{-0.16cm}
			\begin{minipage}{0.1\textwidth}
				\includegraphics[width=\textwidth]{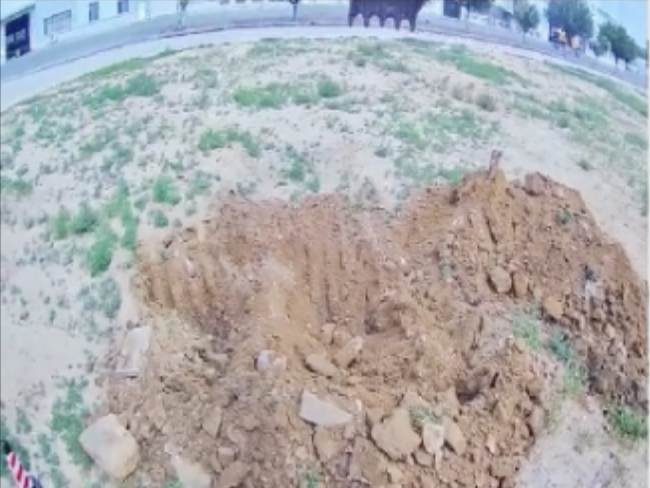}
			\end{minipage}
                \hspace{-0.16cm}
			\begin{minipage}{0.1\textwidth}
				\includegraphics[width=\textwidth]{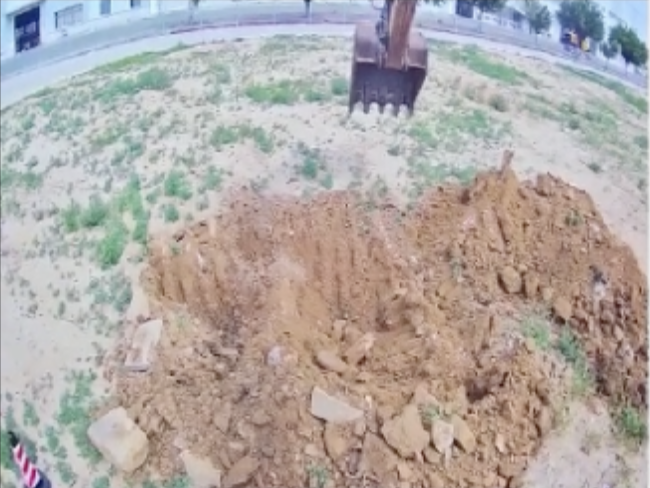}
			\end{minipage}
		}
		
		\vspace{0.1cm}
		{\small(a) Front camera images}
		\vspace{0.1cm}
		
		\scalebox{0.95}{
			\begin{minipage}{0.1\textwidth}
				\includegraphics[width=\textwidth]{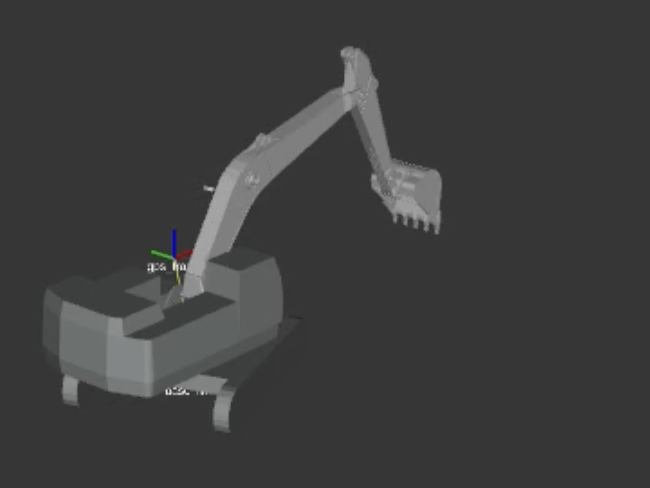}
			\end{minipage}
                \hspace{-0.16cm}
			\begin{minipage}{0.1\textwidth}
				\includegraphics[width=\textwidth]{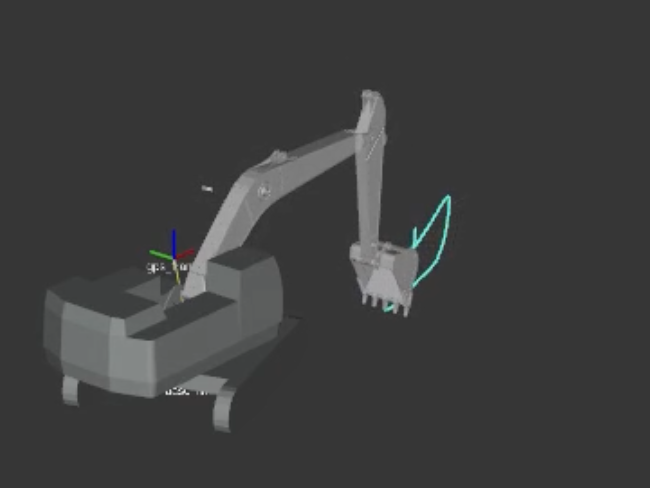}
			\end{minipage}
                \hspace{-0.16cm}
			\begin{minipage}{0.1\textwidth}
				\includegraphics[width=\textwidth]{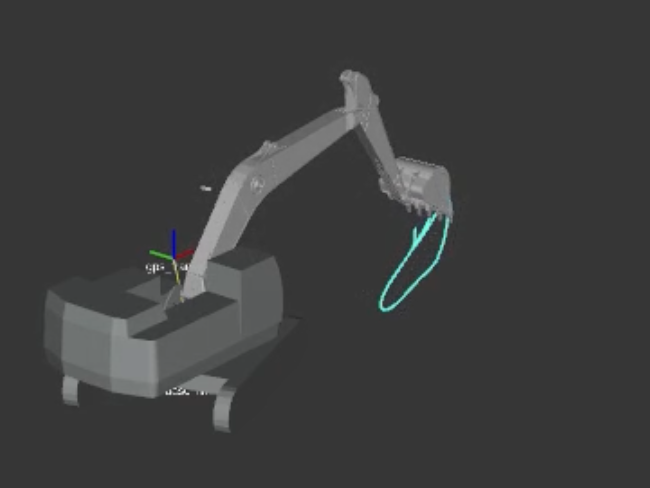}
			\end{minipage}
                \hspace{-0.16cm}
			\begin{minipage}{0.1\textwidth}
				\includegraphics[width=\textwidth]{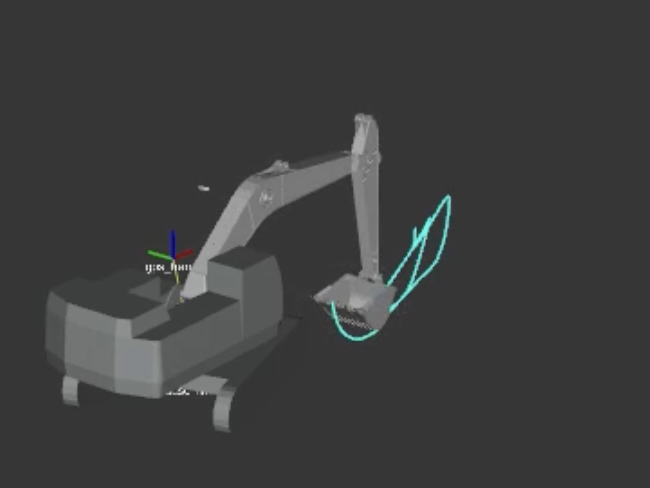}
			\end{minipage}
                \hspace{-0.16cm}
			\begin{minipage}{0.1\textwidth}
				\includegraphics[width=\textwidth]{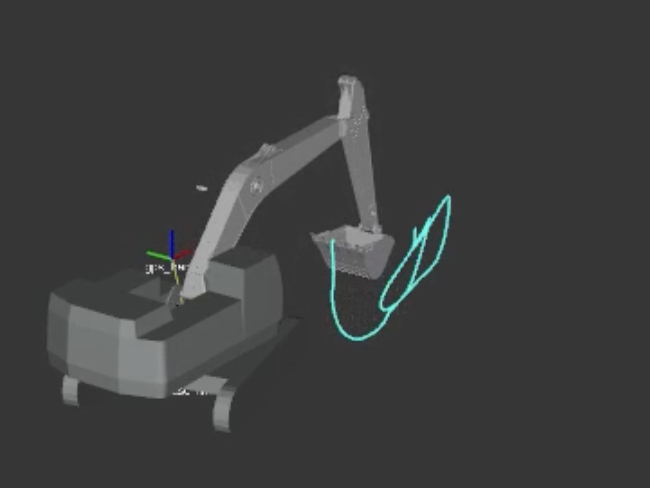}
			\end{minipage}
                \hspace{-0.16cm}
			\begin{minipage}{0.1\textwidth}
				\includegraphics[width=\textwidth]{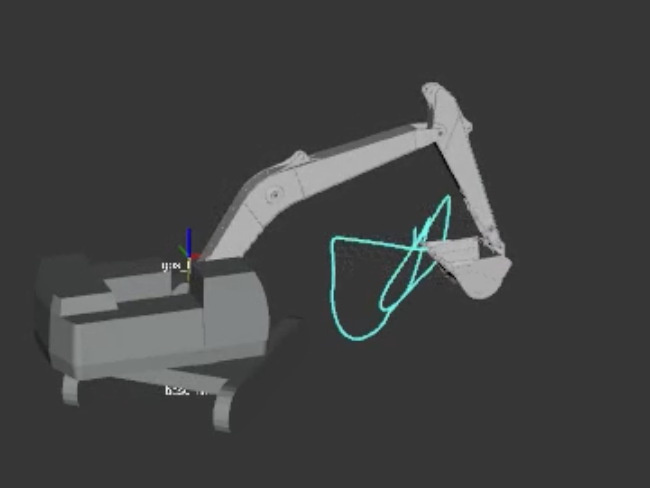}
			\end{minipage}
                \hspace{-0.16cm}
			\begin{minipage}{0.1\textwidth}
				\includegraphics[width=\textwidth]{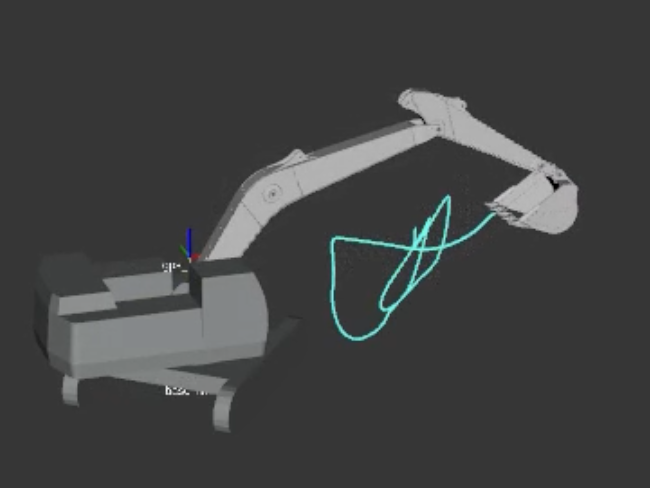}
			\end{minipage}
                \hspace{-0.16cm}
			\begin{minipage}{0.1\textwidth}
				\includegraphics[width=\textwidth]{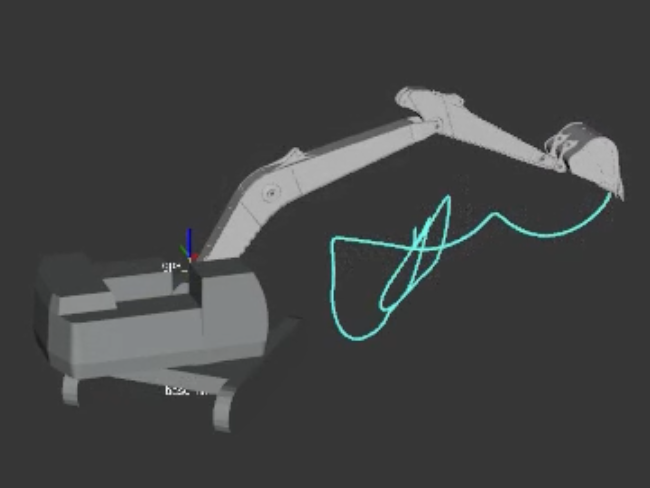}
			\end{minipage}
                \hspace{-0.16cm}
			\begin{minipage}{0.1\textwidth}
				\includegraphics[width=\textwidth]{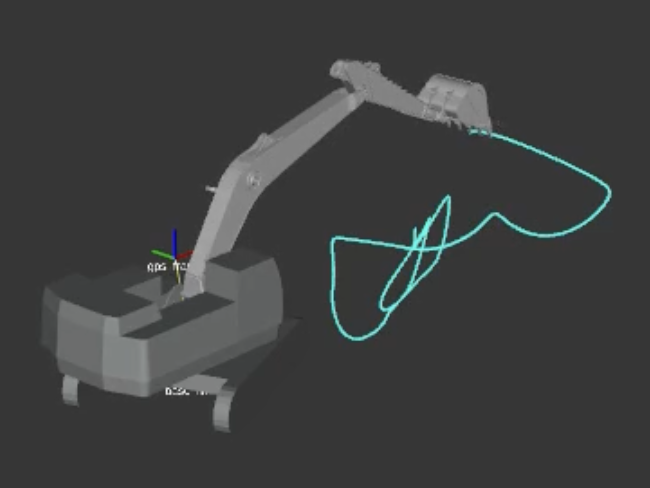}
			\end{minipage}
                \hspace{-0.16cm}
			\begin{minipage}{0.1\textwidth}
				\includegraphics[width=\textwidth]{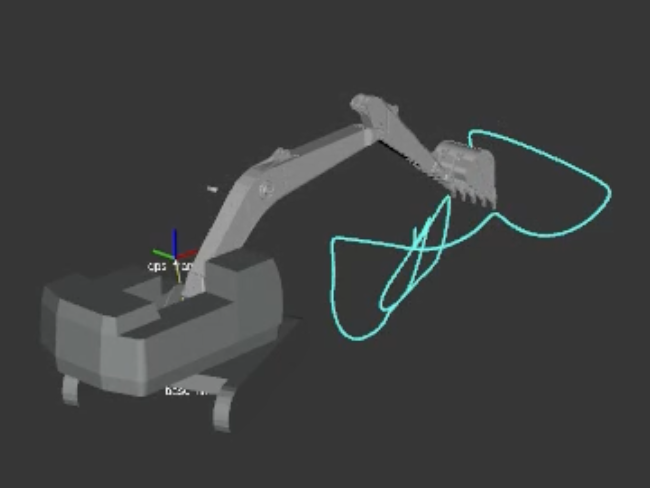}
			\end{minipage}
		}
		
		\vspace{0.1cm}
		{\small(b) Bucket trajectory visualizations in RViz}
		
		\caption{Front camera images and bucket trajectory during the testing of the task \texttt{dig\_dump\_return} (joint position control)}
		\label{Fig2}
	\end{figure*}
	
	\subsection{Data Collections}
	\label{S32}
	We control the valve states of the excavator remotely, which determines the joint velocities, although with the time delay and dead zone effects. The data frequency is 10 Hz. TABLE \ref{T1} shows the statistics of the collected data, including the number of episodes and average episode length in terms of timesteps. In all tasks, we randomly select one episode as a test episode and use all the other episodes as training data. 
	
	\subsection{Valve States to Joint Velocities Modelling}
	\label{S22}
	In our excavator experiment, four valve states are controlled directly to adjust the 4 joint velocities, i.e., swing, boom, stick, and bucket. However, it is unsafe to directly test ExACT on the real excavator without any offline pretest. In our experiments, this offline pretest is achieved by converting the valve states to the corresponding joint velocities via an approximate linear model. Based on the 2661 data of valve states-joint velocities pairs, the linear models are derived as
	\begin{subequations}
		\label{Eq1}
		\begin{align}
			q^{\mathrm{d}}_{\mathrm{swing}}& = -2.8227\times10^{-6} \cdot V_{\mathrm{swing}}+2.3118\times10^{-2},\\
			q^{\mathrm{d}}_{\mathrm{boom}}& = 1.3736\times10^{-6} \cdot V_{\mathrm{boom}}-1.1250\times10^{-2},\\
			q^{\mathrm{d}}_{\mathrm{stick}}& = -2.4656\times10^{-6} \cdot V_{\mathrm{stick}}+2.0193\times10^{-2},\\
			q^{\mathrm{d}}_{\mathrm{bucket}}& = 5.8151\times10^{-6} \cdot V_{\mathrm{bucket}}-4.7625\times10^{-2},
		\end{align}
	\end{subequations}
	where $q^{\mathrm{d}}_{j}$ and $V_{j}$, with $j\in\{\mathrm{swing},\mathrm{boom},\mathrm{stick},\mathrm{bucket}\}$, denote the joint velocities and valve states of the corresponding joints, respectively. Note that the default value of all valve states is 8190, under which the corresponding joint velocity is 0. 
	
	\subsection{Results}
	With the task and data descriptions in Sections \ref{S31} and \ref{S32}, we train ExACT with the hyperparameters in TABLE \ref{T2}. In the test phase, we initialize the joint positions of the first step as the true values in the test episode. The remaining joint positions are calculated based on the corresponding previous joint states and actions. This strategy effectively avoids the phenomenon that the joint positions output by ExACT track the previous real joint positions. For the front camera image and LiDAR elevation maps, we use the real collected data from the test episode. The temporal ensembling method \cite{zhao2023learning} is utilized during the test in all tasks to smooth the trajectory. 

        \begin{table}[!t]
		\centering
		\captionsetup{justification=centering}
		\caption{Data statistics of different tasks}
		\label{T1}
		\begin{tabular}{cwc{2cm}c}
			\hline
			Task & Number of episodes & Average episode length\\
			\hline
			\texttt{reach} & 8 & 180.5 timesteps\\ 
			\texttt{dig\_dump} & 12 & 361.0 timesteps\\ 
			\texttt{dig\_dump\_return} & 12 & 397.8 timesteps\\
			\hline
		\end{tabular}
	\end{table}
	
	\begin{table}[!t]
		\centering
		\captionsetup{justification=centering}
		\caption{Training hyperparameters}
		\label{T2}
		\begin{tabular}{wc{2.5cm}wc{2.5cm}}
			\hline
			Hyperparameter & Value \\
			\hline
			KL divergence weight & 10 \\ 
			Chunk size & 30 \\ 
			Number of steps & 30000 \\
			Learning rate & $1.0\times10^{-5}$ \\
			\hline
		\end{tabular}
	\end{table}
	
	\subsubsection{Results of the \texttt{reach} Task}
	After training ExACT with 8 episodes of demonstration data, we obtain the test results shown in Fig. \ref{Fig1}. In the data collection process, we first align the bucket vertically with the traffic cone and then move down to the target position. Figs. \ref{Fig1}(a)-(b) demonstrate that ExACT learns the correct reaching behavior, and the bucket first moves horizontally and then vertically. The success of the \texttt{reach} task also indicates the linear models of valve states and joint velocities, as shown in Section \ref{S22}, are effective when the excavator task is straightforward. Fig. \ref{Fig1}(c) compares the ground truth and predicted valve states. We note that the predicted value states align well with the ground truths, which further validates the excellent performance of ExACT on the excavator \texttt{reach} task. 
	
	\subsubsection{Results of the \texttt{dig\_dump} Task}
	The ExACT test results of the \texttt{dig\_dump} task after training with 11 episodes are shown in Fig. \ref{Fig6}(a), from which we can find that ExACT struggles to learn the high-frequency components of the valve states. Since the valve states directly impact the joint velocities, the failure to learn the high-frequency components of the valve states will lead to inflexible joint movements. This may cause the bucket edge to move longer than it should during the digging and dumping process in practice. We leave the development of an ExACT controller that can learn the high-frequency components for future work. Moreover, more demonstration data might be collected to improve the ExACT performance on the \texttt{dig\_dump} task with a valve state controller. 
	
	\subsubsection{Results of the \texttt{dig\_dump\_return} Task}
	Due to the imperfect performance of valve state control in the \texttt{dig\_dump} task, we use the joint position control in the \texttt{dig\_dump\_return} task, which includes one additional phase of \texttt{return}. The task visualizations and action predictions during the test are shown in Figs. \ref{Fig6}(b) and \ref{Fig2}, respectively. The results demonstrate that the excavator can complete the whole task perfectly, with the joint positions being predicted with high precision. 
	
	\section{Conclusions}	
	This paper presents ExACT, an end-to-end autonomous system that leverages raw data from LiDAR and cameras and joint positions to control the movements of an excavator. Employing the ACT architecture, this system integrates imitation learning with multi-modal sensor data to produce sequences of actionable commands. Our experimental setup included a simulator developed using linear equations to represent the dynamics between the excavator valve states and its joint velocities. With a limited number of human demonstration trajectories, ExACT successfully executed a variety of excavation tasks in this simulated environment. This achievement marks a significant milestone in the excavator learning field as, to the best of our knowledge, it is the first instance towards building an imitation learning-based end-to-end excavator controller with minimal human demonstrations. In the future, we plan to test the performance of ExACT on real excavators. Our work paves the way for further research and development in autonomous excavation technologies. 
	
	\addtolength{\textheight}{-12cm}   
	
	\bibliographystyle{IEEEtran} 
	\bibliography{bibfile}
	
\end{document}